# Deep Learning for Plankton and Coral Classification

**Alessandra Lumini[1*], Loris Nanni[2], Gianluca Maguolo[2]**

[1] Department of Computer Science and Engineering, University of Bologna, via Pavese 50, 47521, Cesena (FC), Italy.

[2] Department of Information Engineering, University of Padua, via Gradenigo 6/B, 35131 Padova, Italy.

[*]Corresponding Author

emails: alessandra.lumini@unibo.it, loris.nanni@unipd.it, gianluca.maguolo@phd.unipd.it

**Abstract**: In this paper, we present a study about an automated system for monitoring underwater ecosystems. The system here proposed is based on the fusion of different deep learning methods. We study how to create an ensemble based of different Convolutional Neural Network (CNN) models, fine-tuned on several datasets with the aim of exploiting their diversity. The aim of our study is to experiment the possibility of fine-tuning CNNs for underwater imagery analysis, the opportunity of using different datasets for pre-training models, the possibility to design an ensemble using the same architecture with small variations in the training procedure.

Our experiments, performed on 5 well-known datasets (3 plankton and 2 coral datasets) show that the combination of such different CNN models in a heterogeneous ensemble grants a substantial performance improvement with respect to other state-of-the-art approaches in all the tested problems. One of the main contributions of this work is a wide experimental evaluation of famous CNN architectures to report the performance of both the single CNN and the ensemble of CNNs in different problems. Moreover, we show how to create an ensemble which improves the performance of the best single model. The MATLAB source code is freely available at: link provided in title page.

## 1. Introduction

Oceans are the essential lifeblood of the Earth: they provide over 70% of the oxygen and over 97% of the water. Without our oceans, all life, including humans, would not survive. Increases in human population and their resource use have drastically intensified pressures on marine ecosystem services, therefore monitoring and maintaining the oceanic ecosystem is essential to the maintenance of marine habitats. These habitats include plankton population and coral reefs, which are critical to marine food cycles, habitat provision and nutrient cycling [1]. Planktons are one of the main components of ocean ecosystems, due to their function in the oceans food chain. Studying variations of plankton distribution gives useful indicators for oceanic state of health. Coral reefs are among the oldest ecosystems on Earth. They are created by the accumulation of hard calcium carbonate skeletons that coral species leave behind when they die. Not only are coral reef biologically rich ecosystems and a source of natural beauty, they also provide spawning and nursery grounds to many fish populations, protect coastal communities from storm surges and erosion from waves, and give many other services that could be lost forever if a coral reef was degraded or destroyed.

Therefore, the study of plankton and coral distribution is crucial to protect marine ecosystems. In the last years there has been a massive proliferation of digital imagery [2] for the monitoring of underwater ecosystems. Considering that, typically, less than 2% of the acquired imagery can be manually observed by a marine expert, this increase in image data has driven the need for automatic detection and classification systems. Many researchers explored automated methods for performing accurate automatic annotation of marine imagery using computer vision and machine learning based techniques [3]: the accuracy of these systems often depends on the availability of high-quality ground truth dataset.

Deep learning has been certainly one of the most used techniques for underwater imagery analysis within the recent past [3] and a growing number of works use CNN for underwater marine object detection and recognition [4][5]. Researchers have increasingly replaced traditional techniques [6][7], where feature extraction was based on hand-crafted descriptors (such as SIFT and LBP) and classification was done with Support Vector Machines or Random Forests, in favor of deep learning approaches [8][9], that exploit Convolutional Neural Networks (CNN) [10] for image classification. CNNs are multi-layered neural networks whose architecture is somewhat similar to that of the human visual system:

they use restricted receptive fields, and a hierarchy of layers which progressively extract more and more abstracted features. A great advantage of CNNs vs traditional approaches is the use of view-invariant representations learnt from large-scale data which make useless any kind of pre-processing.

The oldest attempts to use deep learning on underwater imagery analysis date back to 2015 in the National Data Science Bowl[1] for plankton image classification. The winner of the competition [11] proposed an ensemble of over 40 convolutional neural networks including layers designed to increase the network robustness to cyclic variation: this is a valid proof of the performance advantage of CNN ensembles vs. single models.

The availability of a large training set has encouraged other works: Py et al. [12] proposed a CNN inspired to GoogleNet improved with an inception module; Lee et al. [8] addressed the class-imbalance problem by performing transfer learning pre-training the CNN on class-normalized data; Dai et al. [9] suggested an ad-hoc model, named ZooplanktoNet inspired by AlexNet and VGGNet; Dai et al. [13] proposed a hybrid 3-channel CNN which takes as input the original image and two preprocessed version of it. When large training sets were not available, automatically labelling was proposed [14] based on Deep Active Learning. Cui et al. [15] proposed a transfer learning approach starting from a model trained on several datasets. In [16] the authors showed that deep learning outperforms handcrafted features for plankton classification and the use of handcrafted approaches is useless also in an ensemble with deep learned methods. In [17] the CNNs are used as feature extractors in combination with a Support Vector Machine for plankton classification.

There are even fewer works that use CNNs for coral classification, since it is a very challenging task due to the high intra-class variance and the fact that some coral species tend to appear together. In [18] the authors proposed a new handcrafted descriptor for coral classification and used a CNN only for the classification step. In [19] Mahmood et al. reported the application of generic CNN representations combined with hand-crafted features for coral reef classification. Afterwards the same authors [20] proposed a framework for coral classification, which employs transfer learning from a pre-trained CNN, thus avoiding the problem of small training set. Beijbom et al. [21] proposed the first architecture specifically designed for coral classification: a five-channel CNN based on CIFAR architecture. Gomez-Rios et al. [22] tested several CNN architectures and transfer learning approaches for classifying coral image from small datasets.

In this work we study ensembles of different CNN models, fine-tuned on several datasets with the aim of exploiting their diversity in designing an ensemble of classifiers. We deal with: *(i)* the ability of fine-tuning pre-trained CNN for underwater imagery analysis, *(ii)* the possibility of using different datasets for pre-training models *(iii)* the possibility of design an ensemble using the same architecture with small variations.

Our ensembles are validated using five well-known datasets (three plankton datasets and two coral datasets) and compared with other state-of-the-art approaches proposed in the literature. Our ensembles based on the combination of different CNNs grant a substantial performance improvement with respect to the state-of-the-art results in all the tested problems. Despite of the complexity in terms of memory requirements, the proposed system has the great benefit of working well "out-of-the-box" in different problems, requiring few parameters tuning without specific pre-processing or optimization for each dataset.

The paper is organized as follows. In section 2 we present the different CNN architectures used in this work, as well as the training options/methods used for fine-tuning the networks. In Section 3, we describe the experimental environments, including the five datasets used for experiments, the testing protocols and the performance indicators; moreover, we suggest and discuss a set of experiments to evaluate our ensembles. In section 4 the conclusion is given along with some proposal for future research.

## *2. Methods*

In this work the deep learned methods are based on fine-tuning well-known CNN architectures according to different training strategies: one and two round training (see the end of this section for details), different activation functions, preprocessing before training. We test several CNN architectures among the most promising models proposed in the literature; the aim of our experiments is both evaluating the most suitable model for these classification problems and considering their diversity to design an ensemble.

---

[1] https://www.kaggle.com/c/datasciencebowl

CNNs are a class of deep neural networks designed for computer vision and image classification, image clustering by similarity, and object recognition. Among the different application of CNNs there are face identification, object recognition, medical image analysis, pedestrian and traffic signs recognition. CNNs are designed to work similarly to the human brain in visually perceiving the world: they are made of neurons (the basic computation units of neural networks), that are activated by specific signals. The neurons of a CNN are stacked in lines called "layers", which are the building blocks of a neural network. A CNN is a repeated concatenation of some classes of (hidden) layers included between the input and output layers [23]:

- Convolutional layers (CONV) perform feature extraction: a CONV layer makes use of a set of learnable filters to detect the presence of specific features or patterns in the input image. Each filter, a matrix with a smaller dimension but the same depth as the input file, is convolved across the input file to return an activation map.
- Activation layers (ACT), implement functions that help to decide if the neuron would fire or not. An activation function is a non linear transformation of the input signal. Since activation functions play a vital role in the training of CNN, several activation functions have been proposed, including Sigmoid, Tanh and Rectified Linear Unit (ReLU). In this work we test a variation of the standard ReLU recently proposed in [24] and named Scaled Exponential Linear Unit (SELU). SELU is basically an exponential function multiplied by an additional parameter, designed to avoid the problem of gradient vanishing or explosion.
- Pooling layers (POOL) are subsampling layers used to reduce the number of parameters and computations in the network with the aim of controlling overfitting. The most used pooling functions are max, average and sum.
- Fully connected layers (FC) are the ones where the neurons are connected to all the activations from the previous layer. The aim of a FC layer is to use the activations from the previous layers for classifying the input image into various classes. Usually the last FC layer basically takes an input volume and outputs an N dimensional vector, where N is the number of classes of the target problem.
- Classification layers (CLASS) perform the final classification selecting the most likely class. They are usually implemented using a SoftMax function in case of single label problem or using a sigmoid activation function with a multiclass output-layer for multi label problems.

In our experiments, we test and combine the following different pre-trained models available in the MATLAB Deep Learning Toolbox; all the models are modified changing the last FC and CLASS layers to fit the number of classes of the target problem, without freezing the weights of the previous layers. Moreover, a variant of each model is evaluated implementing a SELU activation function instead of each ReLU layer that follows a convolution. The models evaluated are:

- AlexNet [25]. AlexNet (the winner of the ImageNet ILSVRC challenge in 2012) is a model including 5 CONV layers followed by 3 FC layers, with some max-POOL layers in the middle. Fast training is achieved applying ReLU activations after each convolutional and fully connected layer. AlexNet accepts images of 227×227 pixels.
- GoogleNet [26]. GoogleNet (the winner of the ImageNet ILSVRC challenge in 2014) is an evolution of AlexNet based on new "Inception" layers (INC), that are a combination of some CONV layers at different granularity, whose outputs are concatenated into a single output vector. This solution makes the network deeper limiting the number of parameters to be inferred. GoogleNet is composed by 27 layers, but has less parameters than AlexNet. GoogleNet accepts input images of 224×224 pixels.
- InceptionV3 [27]. InceptionV3 is an evolution of GoogleNet (also known as Inception1) based on the factorization of 7x7 convolutions into 2 or 3 consecutive layers of 3×3 convolutions. InceptionV3 accepts larger images of 299×299 pixels.
- VGGNet [28]. VGGNet (the network placed second in ILSVRC 2014) is a very deep network which includes 16 or more CONV/FC layers, each based on small 3×3 convolution filters, interspersed by POOL layers (one for each group of 2 or 3 CONV layers). The total number of trainable layers is 23 or more depending on the net: in our experiments we consider two of the best-performing VGG models: VGG-16 and VGG-19, where 16 and 19 stand for the number of layers. The VGG models accept images of 224×224 pixels.
- ResNet [29]. ResNet (the winner of ILSVRC 2015) is a network about 8 times deeper than VGGNet. ResNet introduces a new "network-in-network" architecture using residual (RES) layers. Moreover, differently from above models, ResNet proposes global average pooling layers instead of FC layers at the end of the network. The result is a model deeper than VGGNet, with a smaller size. In this work we use ResNet50 (a 50 layer Residual Network) and ResNet101 (a deeper variant of ResNet50). Both models accept images of 224×224 pixels.
- DenseNet [30]. DenseNet is an evolution of ResNet which includes dense connections among layers: each layer is connected to each following layer in a feed-forward fashion. Therefore the number of connections increases from the number of layers L to L×(L+1)/2. DenseNet improves the performance of previous models at the cost of an augmented computation requirement. DenseNet accepts images of 224×224 pixels.

- MobileNetV2 [31]. MobileNet is a light architecture designed for mobile and embedded vision applications. The model is based on a streamlined architecture that uses depth-wise separable convolutions to build light weight deep neural networks. The network is made of only 54 layers and has an image input size of 224×224.
- NasNet [32]. NasNet is a well performing model, whose architecture is predefined, but blocks or cells are learned by reinforcement learning search method. The basic idea of NasNet is to learn architectural blocks from a small dataset and transfer them on the target problem. The network training is quite heavy and requires large images (input size of 331×331).

In this work we tested three different approaches for fine-tuning the models using one or two training sets of the target problem:

- One round tuning (1R): one round is the standard approach for fine tuning pre-trained networks; the net is initialized according pre-trained weights (obtained on the large ImageNet dataset) and retrained using the training set of the target problem. Differently from other works that fix weights of the first layers, we retrain all layers' weights using the same learning rate in all the network.
- Two rounds tuning (2R): this strategy involves a first round of fine-tuning in a dataset similar to the target one and a second round using the training set of the target problem. The first step consists in fine-tuning the net (initialized according pre-trained weights) on an external dataset including images from classes not incorporated in the target problem. The second step is a One round tuning performed starting from the network tuned on the external dataset. The motivation behind this method is to firstly teach the network to recognize underwater patterns, which are very different from the images in ImageNet dataset, then the second round is used to adjust the classification weights according to the target problem. The plankton and coral datasets used for preliminary tuning are described in section 3.
- Incremental tuning (INC): one of the most important parameters in training is the number of iterations (epochs) used for training. Due to the possibility of overfitting, increasing the number of iterations does not ensure a performance increase. On the other hand, changing the number of iteration introduces a variability which makes the networks diverse to each other. Our incremental tuning strategy is specifically designed to create ensembles and it is based on selecting networks at different training epochs to be combined together in an ensemble. In this work we perform an incremental training with steps of 3 epochs extracting 15 networks.

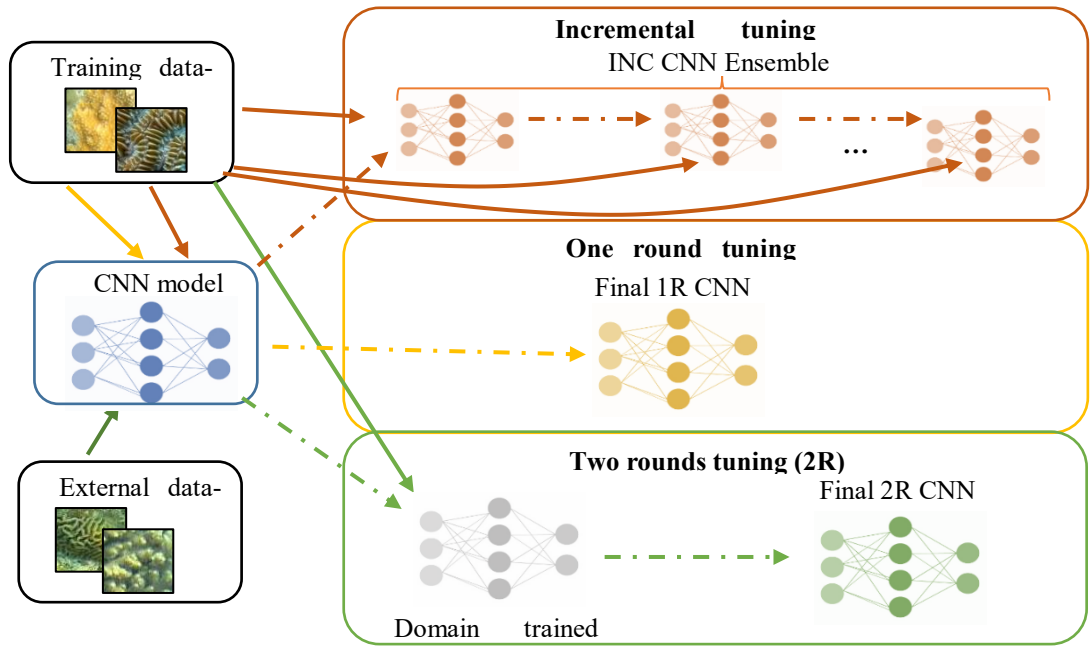

**Fig. 1.** A schema of the three fine-tuning approaches: filled arrows denote input for training, dotted arrows denotes output flows (trained models). Each color is related to a different approach. Yellow arrows are related to 1R tuning, green arrows to 2R tuning, orange ones to INC tuning.

In Fig. 1 a schema of the three methods is reported, where each color is related to a separate approach. One round tuning involves a single fine tuning: the training set is used to fine-tune the input model (yellow arrow) and obtain as output the final 1R tuned CNN (yellow dotted arrow). Two round tuning involves a first tuning of the input model using an external dataset (green arrow), the resulting "domain trained CNN" (output of the first green dotted arrow) is re-tuned using the training set to obtain the final 2R CNN. Incremental tuning starts as the 1R tuning, then sequential tuning steps using the same training set (orange arrows) are preformed to obtain a final ensemble of CNNs (resulting from dotted orange arrows).

The training options are the following: 30 epochs for training (45 for the INC tuning, at steps of 3), mini-batch size varying from 16 to 128 observations (depending on the memory requirements of the model)[2], learning rate of 0.001. Unlike most of works published in the literature, we do not use data augmentation since it did not grant sensible performance improvements in our experiments.

Each model trained according to the three fine-tuning approaches listed above has been evaluated as stand-alone method and as a component of several ensembles. We tested several selection rules to design ensembles: the first exploits a diversity of architectures and it is the fusion of different models trained using the same approach (Fus_1R, Fus_2R…), the second is the fusion of the best stand-alone model (DenseNet) trained using different approaches (DN_1R+2R, DN_1R+2R+INC,…), the third and the fourth are two trained rules whose aim is selecting the best stand-alone models (named SFFS and WS, respectively). SFFS is based on one of the most performing feature selection approach, i.e. Sequential Forward Floating Selection [33], which has been adapted for selecting the most performing/independent classifiers to be added to the ensemble. In the SFFS method, each model to be included in the final ensemble is selected by adding at each step the model which provides the highest incremental of performance to existing subset of models. Then a backtracking step is performed in order to exclude the worst model from the actual ensemble. Since SFFS requires a training phase, in order to select the best suited models, we perform a leave-one-out-dataset selection. The pseudo-code of SFFS selection using leave-one-out dataset is reported in Fig. 2.

A detailed description of each ensemble tested in this work is given in section 3.

---

**Algorithm 1:** SFFS using leave-one-out-dataset
**Input**: A set of trained models M={$m_1,..m_n$}, 5 datasets split in training D={$d_1,..,d_5$} and testing T={$t_1,..,t_5$}
**for** i = 1 to 5 **do**
    leaveInScores := evaluateScores of each model in M on T(i) = T − $t_i$
    bestEnsemble := SFFS(leaveInScores); *//find the best ensemble using SFFS*
    ensembleScores := evaluateScore of bestEnsemble on $t_i$
    predictions := argmax(ensembleScores);
    accuracy(i) := evaluate accuracy;
**end**
avgAccuracy := avg(accuracy)
**Output**: avgAccuracy

---

**Function** SFFS
**Input**: The set of all the classifiers/scores S
**Initialization:** SS:= Ø RS:=S k:=0 *//initialize to the empty set*
**while** the stop criteria is not true
    Select from RS the best score $s_{max}$
    SS:=SS+$s_{max}$ ; RS:=RS−$s_{max}$ ; k:=k+1
    Select from SS the worst score $s_{min}$
    **if** SS− $s_{min}$ outperform SS **then**
        SS:=SS− $s_{min}$ ; RS:=RS+$s_{max}$ ; k:=k-1
    Evaluate the stop criteria

**Fig. 2.** Pseudo-code of SFFS trained using leave-one-out-dataset

The second trained rule, named WS, is a heuristic rule for weighed selection: WS finds a set of weights for all the classifiers and computes the weighted average of the scores of the classifiers. The selection is performed setting to 0 the weight of some classifier. In order to force WS to assign a positive weight to only few classifiers, the loss function is the sum of the usual crossentropy loss and a regularization term given by $L^{REG}=\Sigma w^\gamma$ where $\gamma < 1$. Since the sum of the weights is constrained to be 1, the regularization loss is minimized when only one classifier has a positive weight. Hence, the algorithm must find a balance between a high accuracy and a small number of classifiers. This balance depends on the value of $\gamma$. The optimization is performed using Stochastic Gradient Descent, and the training is performed according to a leave-one-out-dataset protocol.

---
[2] AlexNet, Vgg16, Vgg19, GoogleNet, MobileNetV2: 128; ResNet50: 32; ResNet101, Inceptionv3: 16, NasNet:8

## 3. Experiments

In order to validate our approaches we perform experiments on five well-known datasets (three plankton datasets and two coral datasets): for plankton classification we use the same three datasets used by [7][3], while for coral classification we use two coral datasets tested in [22][4]

- WHOI is a dataset containing 6600 greyscale images stored in tiff format. The images, acquired by Imaging FlowCytobot from Woods Hole Harbor water, belongs to 22 manually categorized plankton classes with equal number of samples for class. In our experiments, we used the same testing protocol proposed by the authors of [34] based on the splitting of the dataset between training and testing sets of equal size.
- ZooScan is a small dataset of 3771 greyscale images acquired using the Zooscan technology from the Bay of Villefranche-sur-mer. Since images contain artifacts (due to manual segmentation), all the images have been automatically cropped before classification. The images belong to 20 classes with variable number of samples for each class. In this work we use the same testing protocol proposed by [7]: 2-fold cross validation.
- Kaggle is a subset, selected by the authors of [7], of the large dataset acquired by ISIIS technology in the Straits of Florida and used for the National Data Science Bowl 2015 competition. The selected subset includes 14374 greyscale images from 38 classes. The distribution among classes is not uniform, but each class has at least 100 samples. In this work we use the same testing protocol proposed by [7]: 5-fold cross validation.
- EILAT is a coral dataset containing 1123 RGB image patches of size 64×64. The patches are cut out from larger images acquired from coral reefs near Eilat in the Red Sea. The dataset is divided into 8 classes characterized by imbalanced distribution. In this work we use the same testing protocol proposed by [22]: 5-fold cross validation.
- RSMAS is a small coral dataset including 766 RGB image patches of size 256×256. The patches are cut out from larger images acquired by the Rosenstiel School of Marine and Atmospheric Sciences of the University of Miami. These images were taken using different cameras in different places. The dataset is divided into 14 classes characterized by an imbalanced distribution. In this work we use the same testing protocol proposed by [22]: 5-fold cross validation.

For the 2-rounds training we used a further training dataset for the plankton problems, obtained by fusing the images from the dataset used for the National Data Science Bowl and not included in the Kaggle dataset (15962 images from 83 classes) and the dataset "Esmeraldo" (11005 samples, 13 classes) obtained from the Zooscan [35] site[5]. For the coral problems we simply perform the first round training using the coral dataset not used for testing: EILAT for RSMAS and vice versa. In Fig. 3 some sample images (2 images per class) from the five datasets are shown. From top to bottom: WHOI, Zooscan, Kaggle, EILAT and RSMAS.

In all the experiments the class distribution has not been maintained when splitting the dataset between training and testing, in order to better deal with the dataset drift problem, i.e. the variation of distribution between training and test set which often causes performance degradation (e.g. [36]). Moreover, we wish to stress that our experiments have been carried out without ad hoc preprocessing for each dataset.

---

[3] Available from https://github.com/zhenglab/PlanktonMKL/tree/master/Dataset

[4] Available from https://data.mendeley.com/datasets/86y667257h/2

[5] http://www.zooscan.obs-vlfr.fr/article.php3?id_article=115 (training) + http://www.zooscan.obs-vlfr.fr/article.php3?id_article=117 (test)

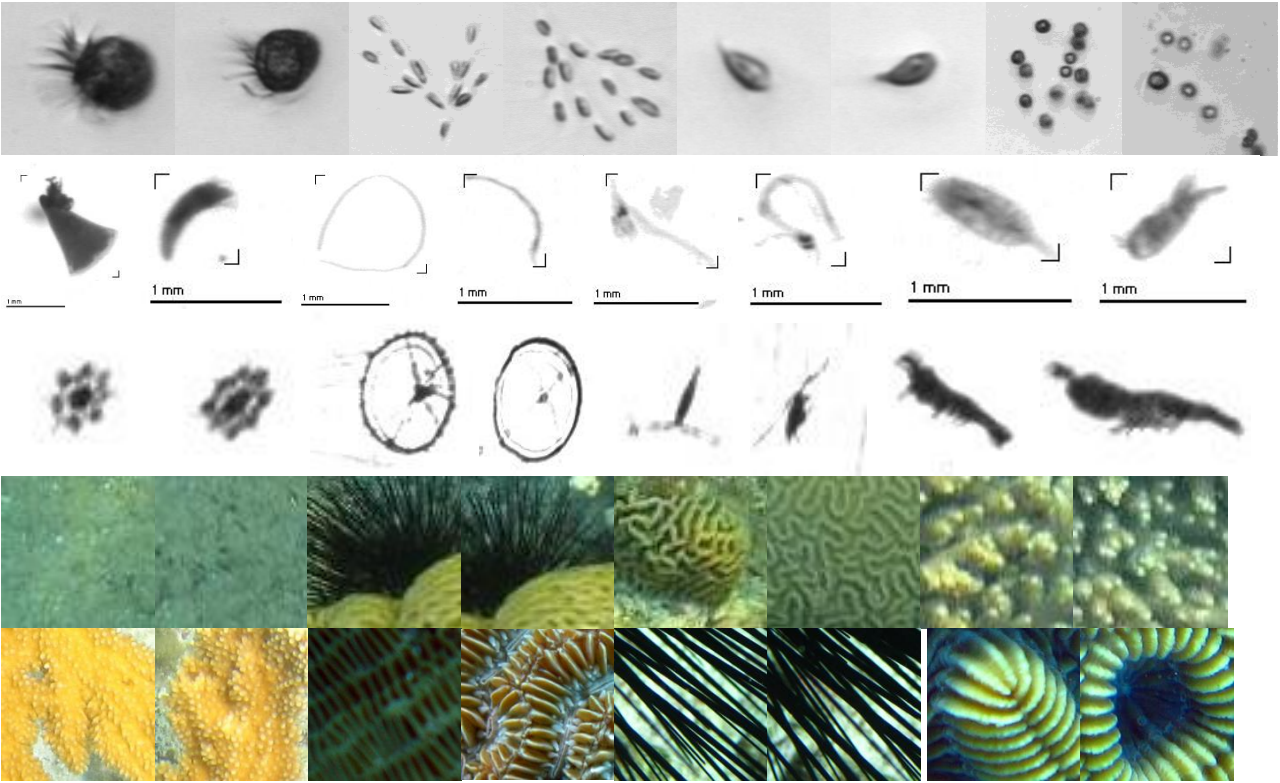

**Fig. 3.** Each row includes 8 sample images from different classes (2 images per class) of the five datasets: WHOI, ZooScan, Kaggle, EILAT, RSMAS.

The evaluation of the proposed approaches and the comparison with the literature is performed according to two of the most used performance indicators in the plankton and coral recognition problems: F-measure and accuracy. In statistical analysis of binary classification, the F-measure (also known as F-score) is a measure of a test's accuracy calculated as the harmonic mean of precision and recall. To extend the definition of F-measure to a multi-class problem the performance indicator is evaluated as the two-class value (one-vs-all) averaged on the number of the classes. Given C confusion matrices $M_c$ related to the C one-vs-all problems, i.e. 2×2 tables including the number of true positive samples ($TP_c$), the number of true negatives ($TN_c$), the number of false positives ($FP_c$) and false negatives ($FN_c$) for each class $c \in [1..C]$, multi-class F-measure is defined as:

- F-Measure is the harmonic mean of precision and recall, $F_C = \frac{P_C \cdot R_C}{P_C + R_C}$ , $F = \frac{1}{C} \sum_C F_C$
- Accuracy is the ratio between the number of true predictions and the total number of samples. $A_C = \frac{TP_C + TN_C}{TP_C + FN_C + FP_C + TN_C}, A = \frac{1}{C} \sum_C A_C.$

The first experiment exhaustively evaluates the ten CNN models according to the One round fine-tuning strategy. Since CNNs require input images at fixed size, we compare 2 different strategies for resizing: square resize (*SqR*) pads the image to square size before resizing to the CNN input size, padding (*Pad*) simply pads the image to the CNN input size (only in few cases where the size of the image is larger than the CNN input size, the image is resized). Padding is performed adding white pixels to plankton images, but it is not suited for RGB coral images, therefore we use tiling (*Tile*) in the 2 coral datasets, consisting in replicating the starting image to a standard size (256×256) and then resizing.

In Table 1 the performance (in terms of F-measure) obtained by different models, fine-tuned according to the 1R strategy, are reported. The results of all the CNNs were obtained using Stochastic Gradient Descent as optimizer, with a fixed learning rate of 0.001. The last two rows in Table 1 report the classification results obtained by the fusion at score level of the above approaches:

- Fus_SqR/ Fus_PT: is the sum rule among the models trained using the same resizing strategy.
- Fus_1R: is the sum rule among Fus_SqR + Fus_PT

DenseNet is the best performing model (Table 1), while NasNet, which has been proved to be one of the most performing architecture in several problems [32], works worse than expected. The reason may be that its automatic block learning is overfitted in ImageNet. Another interesting observation from Table 1 is that the performance of single architectures can be further improved by ensemble approaches. Even the lightweight MobileNetv2, which is one of the worst performing architecture in these datasets, is useful in the ensemble (wrt results reported in [16] where MobileNetV2 was not considered). Since SqR is the resizing strategy that works better in most of the datasets and models, we fixed it for the following experiments. Anyway, it is interesting to note that the fusion among scores obtained from different resizing strategies grants better results than other ensembles.

In tables 2 and 3 exhaustive experiments obtained from different CNN models, using the following methods, are reported (NasNet is excluded for computational reasons): 1R (simple fine tuning using SqR resizing strategy), 2R (2 rounds tuning using SqR resizing strategy), INC (ensemble of models obtained by incremental training), SELU (a variation of each model based on SELU activation, trained by 1R).

**Table 1.** F-measure obtained from different CNN models (1R training), varying the resizing strategy.

| 1R | Dataset | WHOI | | ZooScan | | Kaggle | | Eilat | | RSMAS | |
|---|---|---|---|---|---|---|---|---|---|---|---|
| | Resize Strategy | SqR | Pad | SqR | Pad | SqR | Pad | SqR | Tile | SqR | Tile |
| Model | AlexNet | 0.923 | 0.900 | 0.804 | 0.825 | 0.872 | 0.835 | 0.975 | 0.973 | 0.947 | 0.947 |
| | GoogleNet | 0.935 | 0.931 | 0.836 | 0.841 | 0.890 | 0.869 | 0.978 | **0.981** | 0.974 | 0.967 |
| | InceptionV3 | 0.947 | 0.939 | 0.843 | **0.856** | 0.904 | 0.869 | 0.966 | 0.969 | 0.963 | 0.952 |
| | VGG16 | 0.940 | 0.936 | 0.847 | 0.863 | 0.890 | 0.881 | **0.983** | 0.979 | 0.971 | 0.964 |
| | VGG19 | 0.939 | 0.937 | 0.840 | 0.848 | 0.890 | 0.873 | 0.978 | **0.981** | 0.971 | 0.955 |
| | ResNet50 | 0.939 | 0.929 | 0.847 | 0.834 | 0.898 | 0.871 | 0.967 | **0.981** | 0.970 | 0.965 |
| | ResNet101 | 0.938 | 0.944 | 0.848 | 0.825 | 0.904 | **0.887** | 0.969 | 0.963 | 0.974 | 0.960 |
| | DenseNet | 0.949 | **0.945** | **0.878** | 0.851 | **0.912** | **0.887** | 0.969 | 0.968 | **0.979** | **0.973** |
| | NasNet | **0.950** | 0.943 | 0.861 | 0.834 | 0.904 | **0.887** | 0.939 | 0.954 | 0.944 | 0.948 |
| | MobileNetV2 | 0.927 | 0.931 | 0.819 | 0.807 | 0.886 | 0.859 | 0.950 | 0.952 | 0.942 | 0.947 |
| Ensemble | Fus_SqR/ Fus_PT | **0.953** | 0.950 | **0.888** | 0.886 | **0.925** | 0.912 | 0.986 | 0.989 | 0.989 | 0.988 |
| | Fus_1R | 0.954 | | 0.894 | | 0.924 | | 0.990 | | 0.994 | |

**Table 2.** F-measure obtained from different CNN models using the following methods on the Plankton datasets (VGG16 and VGG19 do not converge using SELU activation function).

| Dataset | WHOI | | | | ZooScan | | | | Kaggle | | | |
|---|---|---|---|---|---|---|---|---|---|---|---|---|
| Method | 1R | 2R | INC | SELU | 1R | 2R | INC | SELU | 1R | 2R | INC | SELU |
| AlexNet | 0.923 | 0.920 | 0.914 | 0.914 | 0.804 | 0.839 | 0.816 | 0.813 | 0.872 | 0.880 | 0.882 | 0.879 |
| GoogleNet | 0.935 | 0.940 | 0.935 | **0.941** | 0.836 | 0.854 | 0.835 | **0.861** | 0.890 | 0.894 | 0.965 | 0.890 |
| InceptionV3 | 0.947 | 0.944 | **0.953** | 0.941 | 0.843 | 0.849 | 0.863 | **0.861** | 0.904 | 0.909 | 0.910 | 0.907 |
| VGG16 | 0.940 | 0.929 | 0.940 | -- | 0.847 | 0.840 | 0.853 | -- | 0.890 | 0.887 | 0.904 | -- |
| VGG19 | 0.939 | 0.930 | 0.933 | -- | 0.840 | 0.831 | 0.846 | -- | 0.890 | 0.871 | **0.914** | -- |
| ResNet50 | 0.939 | 0.932 | 0.936 | 0.928 | 0.847 | 0.863 | 0.841 | 0.847 | 0.898 | 0.903 | 0.908 | 0.903 |
| ResNet101 | 0.938 | 0.938 | 0.941 | 0.937 | 0.848 | 0.869 | 0.837 | 0.843 | 0.904 | 0.904 | 0.909 | **0.909** |
| DenseNet | **0.949** | **0.947** | 0.951 | 0.882 | **0.878** | **0.882** | 0.876 | 0.763 | **0.912** | **0.914** | 0.913 | 0.853 |
| MobileNetV2 | 0.927 | 0.924 | 0.935 | 0.928 | 0.819 | 0.823 | 0.848 | 0.833 | 0.886 | 0.892 | 0.901 | 0.896 |

**Table 3.** F-measure obtained from different CNN models using the following methods on the Coral datasets (VGG16 and VGG19 do not converge using SELU activation function).

| Dataset | EILAT | | | | RSMAS | | | |
|---|---|---|---|---|---|---|---|---|
| Method | 1R | 2R | INC | SELU | 1R | 2R | INC | SELU |
| AlexNet | 0.975 | 0.954 | 0.975 | 0.980 | 0.947 | 0.901 | 0.962 | 0.943 |
| GoogleNet | 0.978 | 0.966 | 0.974 | 0.982 | 0.974 | 0.942 | 0.971 | 0.969 |
| InceptionV3 | 0.966 | 0.971 | 0.968 | 0.972 | 0.963 | 0.954 | 0.961 | 0.969 |
| VGG16 | **0.983** | 0.967 | 0.982 | --- | 0.971 | 0.952 | 0.982 | --- |
| VGG19 | 0.978 | 0.969 | **0.988** | --- | 0.971 | 0.922 | 0.981 | --- |
| ResNet50 | 0.967 | 0.962 | 0.975 | 0.977 | 0.970 | 0.961 | 0.981 | **0.980** |
| ResNet101 | 0.969 | **0.973** | 0.971 | **0.983** | 0.974 | 0.973 | **0.988** | 0.979 |
| DenseNet | 0.969 | 0.972 | 0.985 | 0.951 | **0.979** | **0.974** | 0.983 | 0.930 |
| MobileNetV2 | 0.950 | 0.952 | 0.973 | 0.966 | 0.942 | 0.938 | 0.966 | 0.954 |

In Table 4 the results obtained by several ensembles are reported. We consider both the ensembles obtained fusing all the CNN models trained with the same strategy (named "Fus_*") and the fusion of the best single model DenseNet (named "DN_*") trained according to different approaches:

- Fus_1R is the fusion (already reported in Table 1) among the models trained by 1R tuning
- Fus_2R is the fusion among the models trained by 2R tuning
- Fus_INC is the fusion of the ensembles obtained by incremental training
- Fus_SELU is the fusion among the models modified by means of SELU activation layers.
- DN_1R is the fusion among the two DenseNet models fine-tuned by 1R tuning using two resizing strategies (SqR + Pad/Tile)
- DN_1R+2R is the fusion among DN_1R and the DenseNet model trained by 2R tuning
- DN_1R+2R+INC is the fusion among DN_1R+2R and the INC version of DenseNet
- DN_1R+2R+INC+SELU is the fusion among the above ensemble and the SELU version of DenseNet

The last column of Table 4 shows the Rank of the average rank, which is obtained by ranking methods for each dataset, averaging the results and ranking again the approaches.

**Table 4.** F-measure obtained from different ensembles using the following methods on the five datasets (* VGG models are not considered).

| Dataset | WHOI | ZooScan | Kaggle | EILAT | RSMAS | Rank |
|---|---|---|---|---|---|---|
| Fus_1R | 0.954 | 0.894 | 0.924 | 0.990 | **0.994** | 2 |
| Fus_2R | 0.952 | 0.891 | 0.923 | **0.991** | **0.994** | 5 |
| Fus_INC | **0.956** | 0.886 | **0.935** | 0.985 | 0.989 | 6 |
| Fus_SELU* | 0.943 | 0.869 | 0.922 | 0.987 | 0.987 | 10 |
| Fus_2R + Fus_1R | 0.955 | **0.899** | 0.926 | 0.989 | **0.994** | 1 |
| Fus_SELU+ Fus_1R | 0.951 | 0.892 | 0.925 | 0.989 | **0.994** | 4 |
| DN_1R | 0.953 | 0.880 | 0.917 | 0.976 | 0.991 | 9 |
| DN_1R+2R | 0.955 | 0.894 | 0.924 | 0.980 | **0.994** | 3 |
| DN_1R+2R+INC | 0.954 | 0.880 | 0.916 | 0.981 | 0.991 | 8 |
| DN_1R+2R+INC+SELU | 0.953 | 0.891 | 0.922 | 0.983 | 0.993 | 7 |

From the results in Tables 2 and 3 it is clear that a single fine tuning is enough for the tested problem, maybe because the datasets used in the first round tuning are not sufficiently similar to the target problem or more probably because the dimension of the training set of the target problem is large enough to perform training. The INC version of each model slightly improves the performance in some cases but does not grant a substantial advantage. As to SELU, it works better than ReLU only in few cases and does not work in VGG models. Anyway from the ensembles of Table 4 we can see that the use of a preliminary training (2R) or other variations allows to create classifiers diverse from 1R and their fusion can significantly improve the performance in these classification problems. Clearly the ensembles of different CNN models (Fus_*) strongly outperform the stand-alone CNNs in all the five tested datasets. However, due to computational reasons, we also considered lighter ensembles based on a single architecture (we selected the most performing one, i.e. DenseNet): it is interesting to note that DN_1R+2R obtains a very good performance using only three networks.

Moreover, we make some experiments using a CNN as feature extractor for training Support Vector Machine classifiers. We used the same approach proposed in [37] starting from DenseNet trained by 1R-SqR approach. The results are reported in Table 5: the first row reports the same performance of DenseNet trained by 1R-SqR of table 2 (here named DN_SqR), the second row reports the performance of the ensemble of SVM trained using the features extracted by DenseNet (DN_SVM); the last row Sum is the sum rule between DN_SVM and DN_SqR. Unfortunately, the performance improvement is almost negligible, anyway, a slight improvement is obtained in all the datasets.

**Table 5.** Transfer learning performance (F-measure).

| Dataset | WHOI | ZooScan | Kaggle | EILAT | RSMAS |
|---|---|---|---|---|---|
| DN_SqR | 0.949 | **0.878** | 0.912 | 0.969 | 0.979 |
| DN_SVM | 0.935 | 0.860 | 0.911 | 0.969 | 0.972 |
| DN_SqR+ DN_SVM | **0.951** | **0.878** | **0.914** | **0.970** | **0.980** |

The last experiment is aimed at reducing the computational requirement of the best ensemble. To this aim, we tested two "classifier selection approaches" as detailed in section 2: SFFS and WS.

The results obtained selecting 11 and 3 classifiers are reported in Table 6 and are very interesting, since they demonstrate that a reduced set of 11 classifiers improve the performance with respect to the previous best ensemble. Using less classifiers permits to develop a lighter approach, SFFS(11 classifiers) has an average memory usage (measured as the sum of occupation of the CNNs models) of 2026MB while WS(11 classifiers) has an average memory usage of 2129MB, SFFS(3 classifiers) has an average memory usage of 582.8 MB while WS(3 classifiers) has a memory usage of 502 MB, respect the ~5.5 GB of Fus_2R + Fus_1R.

**Table 6.** F-measure obtained from reduced ensembles.

| Dataset | WHOI | ZooScan | Kaggle | EILAT | RSMAS |
|---|---|---|---|---|---|
| Fus_2R + Fus_1R (27 classifiers) | 0.955 | 0.899 | 0.926 | 0.989 | 0.994 |
| SFFS(11 classifiers) | **0.958** | 0.900 | **0.927** | **0.990** | **0.995** |
| SFFS(3 classifiers) | 0.954 | 0.889 | 0.921 | 0.979 | 0.984 |
| WS(11 classifiers) | **0.958** | 0.902 | **0.927** | 0.987 | 0.993 |
| WS(3 classifiers) | 0.956 | 0.895 | 0.923 | 0.981 | 0.985 |

The performance increase obtained starting from the best stand-alone approach (i.e. DN_SqR) to our best ensemble is shown in Fig. 4 where each step towards our best solution is compared. To show that the performance increase, in case of imbalanced distribution, is not related only to larger classes, but also to small classes we show in Fig. 5 the confusion matrices obtained by DN_SqR and SFFS(11) on the ZooScan dataset which is the dataset with larger performance increase (the confusion matrices on the other datasets are included as supplemental material). The comparison of the two matrices confirms an improvement distributed over all the classes.

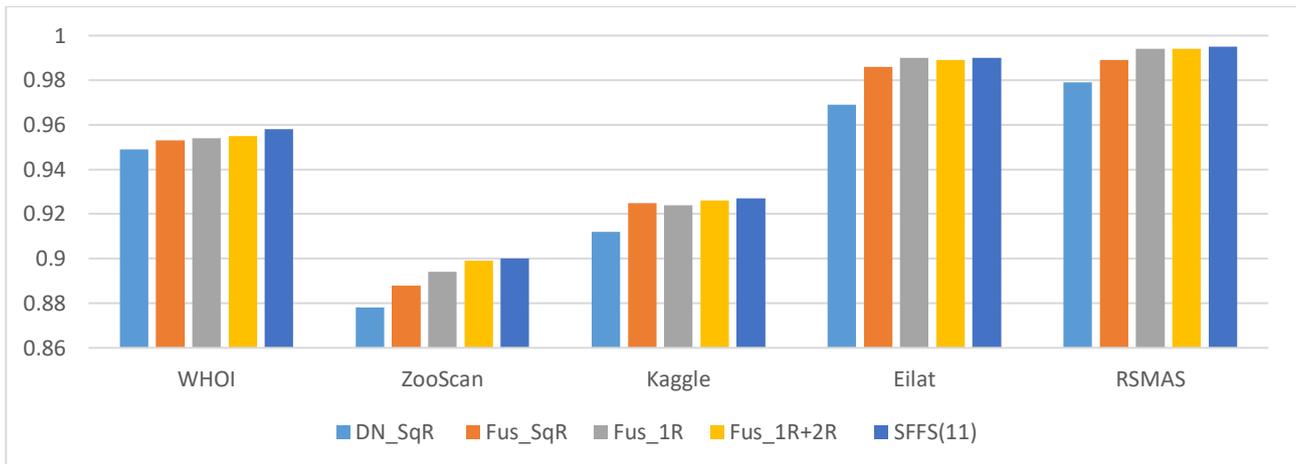

**Fig. 4.** Comparison of different approaches in terms of F-measure on the 5 datasets.

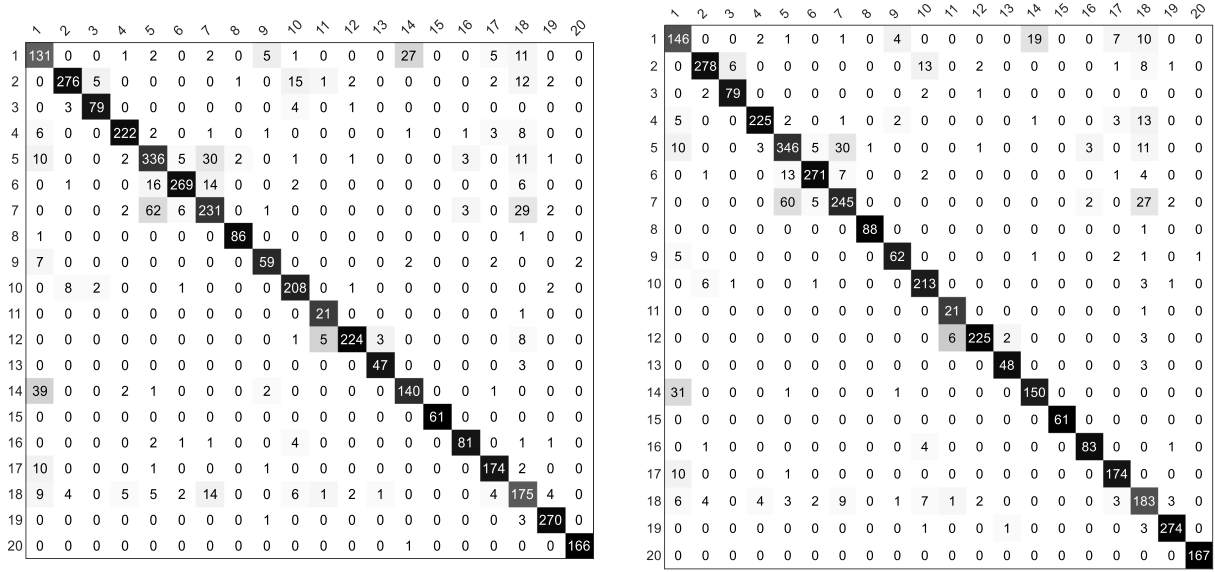

**Fig. 5.** Comparison of the confusion matrices for DN_SqR(left) and SFFS(11) (right) on the ZooScan dataset.

Finally, in Table 7 and Table 8 we report the comparison among the ensembles proposed in this work and other state-of-the-art approaches evaluated on the same datasets:
- FUS_Hand [38] is an ensemble of handcrafted descriptors;
- Gaussian SVM [7] is a handcrafted approach based on a SVM classifier.
- MKL [7] is a handcrafted approach based on multiple kernel learning classifiers.
- DeepL [22] is a deep learned approach based on ResNet
- Opt [18] is handcrafted approach based on a novel feature descriptor. The reported result is the best among all tested feature descriptors.
- EnsHC [39] is an ensemble of several handcrafted features (i.e. completed local binary pattern, grey level co-occurrence matrix, Gabor filter response, …) and classifiers.

The results are reported in terms of F-measures and accuracy depending on the performance indicator used in the literature. The same testing protocol is used in all the methods. The same ensemble, not adapted in each given dataset, obtains state of the art results in all the five tested datasets.

**Table 7.** Comparison vs. state-of-the-art methods (F-measure). * indicates slightly different testing protocol (75%training, 25%test).

| Dataset | WHOI | ZooScan | Kaggle | EILAT | RSMAS |
|---|---|---|---|---|---|
| Fus_2R + Fus_1R | 0.955 | 0.899 | 0.926 | 0.989 | 0.994 |
| DN_1R+2R | 0.955 | 0.894 | 0.924 | 0.980 | 0.994 |
| SFFS (11 classifiers) | **0.958** | 0.900 | **0.927** | **0.990** | **0.995** |
| WS (11 classifiers) | **0.958** | 0.902 | **0.927** | 0.987 | 0.993 |
| FUS_Hand [38] | 0.903 | 0.843 | 0.849 | --- | --- |
| Gaussian SVM [7] | 0.896 | 0.861 | 0.830 | --- | --- |
| MKL (3 kernels) [7] | 0.900 | 0.894 | 0.846 | --- | --- |
| OPT[18]* | | | | 0.88 | 0.863 |

**Table 8.** Comparison vs. state-of-the-art methods (Accuracy). ** indicates a different testing protocol: 10-fold cross validation.

| Dataset | WHOI | ZooScan | Kaggle | EILAT | RSMAS |
|---|---|---|---|---|---|
| Fus_2R + Fus_1R | 95.5 | 88.6 | **94.2** | 98.8 | 99.1 |
| DN_1R+2R | 95.5 | 87.7 | 93.9 | 97.9 | 99.1 |
| SFFS (11 classifiers) | **95.8** | 88.5 | **94.2** | **98.9** | **99.2** |
| WS (11 classifiers) | **95.8** | 88.8 | 94.2 | 98.7 | 99.0 |
| DeepL [22] | --- | --- | --- | 97.85 | 97.95 |
| EnsHC [39] ** | --- | --- | --- | 96.9 | 96.5 |

Finally, we ran SFFS using all the five datasets. In this way we find an ensemble of 11 CNNs that includes:
1. AlexNet_1R_ Pad/Tile
2. AlexNet_2R
3. vgg16_1R_SqR
4. vgg16_INC
5. googlenet_1R_ Pad/Tile
6. googlenet_INC
7. resnet50_2R
8. inceptionv3_1R_ Pad/Tile
9. densenet201_1R_ Pad/Tile
10. densenet201_2R
11. nasnetlarge_1R_SqR

## *4. Conclusions*

Underwater imagery analysis is a challenging task due to the large number of different classes, the great intra-class variance, the low extra-class differences and the lightning variations due to the water. In this paper we studied several deep learned approaches for plankton and coral classification with the aim of exploiting their diversity for designing an ensemble of classifiers. Our final system is based on the fine-tuning of several CNN models trained according to different strategies, which fused together in a final ensemble gain higher performance than the single CNN. In our experiments, carried out on 5 datasets (3 plankton and 2 coral ones), we evaluated well-known CNN models fine-tuned on the target problem using some training variations (different resizing for input images, tuning on similar datasets, small variations of the original CNN model): the experimental results show that the best stand-alone model for most of the target datasets is DenseNet, anyway the combination of several CNNs in an ensemble grants a substantial performance improvement with respect to the single best model.

In order to reduce the complexity of the resulting ensemble, we used a feature selection approach aimed at selecting the best classifiers to be included in the fusion: the final result is a lighter version of the ensemble including only 11 classifiers which outperforms all the other ensembles proposed.

All the MATLAB code used in our experiments will be freely available in our GitHub repository (https://github.com/LorisNanni) in order to reproduce the experiments reported and for future comparisons.

**Acknowledgments** We would like to acknowledge the support that NVIDIA provided us through the GPU Grant Program. We used a donated TitanX GPU to train CNNs used in this work.

*References*

[1] A.W.D. Larkum, R.J. Orth, C.M. Duarte, Seagrasses: Biology, ecology and conservation, 2006. doi:10.1007/978-1-4020-2983-7.

[2] R.J. Olson, H.M. Sosik, A submersible imaging-in-flow instrument to analyze nano-and microplankton: Imaging FlowCytobot, Limnol. Oceanogr. Methods. 5 (2007) 195–203. doi:10.4319/lom.2007.5.195.

[3] M. Moniruzzaman, S.M.S. Islam, M. Bennamoun, P. Lavery, Deep learning on underwater marine object detection: A survey, in: Lect. Notes Comput. Sci. (Including Subser. Lect. Notes Artif. Intell. Lect. Notes Bioinformatics), 2017: pp. 150–160. doi:10.1007/978-3-319-70353-4_13.

[4] H. Qin, X. Li, Z. Yang, M. Shang, When underwater imagery analysis meets deep learning: A solution at the age of big visual data, in: Ocean. 2015 - MTS/IEEE Washingt., 2016.

[5] D. Rathi, S. Jain, S. Indu, Underwater Fish Species Classification using Convolutional Neural Network and Deep Learning, in: 2017 9th Int. Conf. Adv. Pattern Recognition, ICAPR 2017, 2018. doi:10.1109/ICAPR.2017.8593044.

[6] F. Zhao, F. Lin, H.S. Seah, Binary SIPPER plankton image classification using random subspace, Neurocomputing. 73 (2010) 1853–1860. doi:10.1016/j.neucom.2009.12.033.

[7] H. Zheng, R. Wang, Z. Yu, N. Wang, Z. Gu, B. Zheng, Automatic plankton image classification combining multiple view features via multiple kernel learning, BMC Bioinformatics. 18 (2017) 1–18. doi:10.1186/s12859-017-1954-8.

[8] H. Lee, M. Park, J. Kim, Plankton classification on imbalanced large scale database via convolutional neural networks with transfer learning, in: Proc. - Int. Conf. Image Process. ICIP, 2016: pp. 3713–3717. doi:10.1109/ICIP.2016.7533053.

[9] J. Dai, R. Wang, H. Zheng, G. Ji, X. Qiao, ZooplanktoNet: Deep convolutional network for zooplankton classification, in: Ocean. 2016 - Shanghai, 2016. doi:10.1109/OCEANSAP.2016.7485680.

[10] J. Gu, Z. Wang, J. Kuen, L. Ma, A. Shahroudy, B. Shuai, et al., Recent advances in convolutional neural networks, Pattern Recognit. 77 (2018) 354–377. doi:10.1016/J.PATCOG.2017.10.013.

[11] S. Dieleman, J. De Fauw, K. Kavukcuoglu, Exploiting Cyclic Symmetry in Convolutional Neural Networks, CoRR. abs/1602.0 (2016). http://arxiv.org/abs/1602.02660.

[12] O. Py, H. Hong, S. Zhongzhi, Plankton classification with deep convolutional neural networks, in: 2016 IEEE Inf. Technol. Networking, Electron. Autom. Control Conf., 2016: pp. 132–136. doi:10.1109/ITNEC.2016.7560334.

[13] J. Dai, Z. Yu, H. Zheng, B. Zheng, N. Wang, A Hybrid Convolutional Neural Network for Plankton Classification, in: C.-S. Chen, J. Lu, K.-K. Ma (Eds.), Comput. Vis. -- ACCV 2016 Work., Springer International Publishing, Cham, 2017: pp. 102–114.

[14] E. Bochinski, G. Bacha, V. Eiselein, T.J.W. Walles, J.C. Nejstgaard, T. Sikora, Deep Active Learning for In Situ Plankton Classification, in: Z. Zhang, D. Suter, Y. Tian, A. Branzan Albu, N. Sidère, H. Jair Escalante (Eds.), Pattern Recognit. Inf. Forensics, Springer International Publishing, Cham, 2019: pp. 5–15.

[15] F.C.M. Rodrigues, N.S.T. Hirata, A.A. Abello, L.T.D. La Cruz, R.M. Lopes, R.H. Jr., Evaluation of Transfer Learning Scenarios in Plankton Image Classification, in: Proc. 13th Int. Jt. Conf. Comput. Vision, Imaging

Comput. Graph. Theory Appl. - Vol. 5 VISAPP, SciTePress, 2018: pp. 359–366. doi:10.5220/0006626703590366.

[16] A. Lumini, L. Nanni, Deep learning and transfer learning features for plankton classification, Ecol. Inform. (2019). doi:10.1016/j.ecoinf.2019.02.007.

[17] K. Cheng, X. Cheng, Y. Wang, H. Bi, M.C. Benfield, Enhanced convolutional neural network for plankton identification and enumeration, PLoS One. (2019). doi:10.1371/journal.pone.0219570.

[18] A.B.M. N, D. Dharma, Coral reef image/video classification employing novel octa-angled pattern for triangular sub region and pulse coupled convolutional neural network (PCCNN), Multimed. Tools Appl. 77 (2018) 31545–31579. doi:10.1007/s11042-018-6148-5.

[19] A. Mahmood, M. Bennamoun, S. An, F. Sohel, F. Boussaid, R. Hovey, et al., Coral classification with hybrid feature representations, in: Proc. - Int. Conf. Image Process. ICIP, 2016. doi:10.1109/ICIP.2016.7532411.

[20] A. Mahmood, M. Bennamoun, S. An, F.A. Sohel, F. Boussaid, R. Hovey, et al., Deep Image Representations for Coral Image Classification, IEEE J. Ocean. Eng. (2019). doi:10.1109/JOE.2017.2786878.

[21] O. Beijbom, T. Treibitz, D.I. Kline, G. Eyal, A. Khen, B. Neal, et al., Improving Automated Annotation of Benthic Survey Images Using Wide-band Fluorescence, Sci. Rep. (2016). doi:10.1038/srep23166.

[22] A. Gómez-Ríos, S. Tabik, J. Luengo, A.S.M. Shihavuddin, B. Krawczyk, F. Herrera, Towards highly accurate coral texture images classification using deep convolutional neural networks and data augmentation, Expert Syst. Appl. (2019). doi:10.1016/j.eswa.2018.10.010.

[23] A. Goodfellow, Ian, Bengio, Yoshua, Courville, Deep Learning, MIT Press. (2016). http://www.deeplearningbook.org/.

[24] G. Klambauer, T. Unterthiner, A. Mayr, S. Hochreiter, Self-Normalizing Neural Networks, in: NIPS, 2017. http://arxiv.org/abs/1706.02515 (accessed June 19, 2019).

[25] A. Krizhevsky, I. Sutskever, G.E. Hinton, ImageNet Classification with Deep Convolutional Neural Networks, Adv. Neural Inf. Process. Syst. (2012) 1–9. doi:http://dx.doi.org/10.1016/j.protcy.2014.09.007.

[26] C. Szegedy, W. Liu, Y. Jia, P. Sermanet, S. Reed, D. Anguelov, et al., Going deeper with convolutions, in: Proc. IEEE Comput. Soc. Conf. Comput. Vis. Pattern Recognit., 2015: pp. 1–9. doi:10.1109/CVPR.2015.7298594.

[27] C. Szegedy, V. Vanhoucke, S. Ioffe, J. Shlens, Z. Wojna, Rethinking the Inception Architecture for Computer Vision, in: 2016 IEEE Conf. Comput. Vis. Pattern Recognit., 2016: pp. 2818–2826. doi:10.1109/CVPR.2016.308.

[28] K. Simonyan, A. Zisserman, Very Deep Convolutional Networks for Large-Scale Image Recognition, Int. Conf. Learn. Represent. (2015) 1–14. doi:10.1016/j.infsof.2008.09.005.

[29] K. He, X. Zhang, S. Ren, J. Sun, Deep Residual Learning for Image Recognition, in: 2016 IEEE Conf. Comput. Vis. Pattern Recognit., 2016: pp. 770–778. doi:10.1109/CVPR.2016.90.

[30] G. Huang, Z. Liu, L. Van Der Maaten, K.Q. Weinberger, Densely connected convolutional networks, in: Proc. - 30th IEEE Conf. Comput. Vis. Pattern Recognition, CVPR 2017, 2017. doi:10.1109/CVPR.2017.243.

[31] M. Sandler, A. Howard, M. Zhu, A. Zhmoginov, L.C. Chen, MobileNetV2: Inverted Residuals and Linear Bottlenecks, in: Proc. IEEE Comput. Soc. Conf. Comput. Vis. Pattern Recognit., 2018. doi:10.1109/CVPR.2018.00474.

[32] B. Zoph, V. Vasudevan, J. Shlens, Q. V. Le, Learning Transferable Architectures for Scalable Image Recognition, in: Proc. IEEE Comput. Soc. Conf. Comput. Vis. Pattern Recognit., 2018. doi:10.1109/CVPR.2018.00907.

[33] P. Pudil, J. Novovičová, J. Kittler, Floating search methods in feature selection, Pattern Recognit. Lett. (1994). doi:10.1016/0167-8655(94)90127-9.

[34] H.M. Sosik, R.J. Olson, Automated taxonomic classification of phytoplankton sampled with imaging-in-flow cytometry, Limnol. Oceanogr. Methods. 5 (2007) 204–216. doi:10.4319/lom.2007.5.204.

[35] G. Gorsky, M.D. Ohman, M. Picheral, S. Gasparini, L. Stemmann, J.B. Romagnan, et al., Digital zooplankton image analysis using the ZooScan integrated system, J. Plankton Res. 32 (2010) 285–303. doi:10.1093/plankt/fbp124.

[36] P. Gonzalez, E. Alvarez, J. Diez, A. Lopez-Urrutia, J.J. del Coz, Validation methods for plankton image classification systems, Limnol. Oceanogr. Methods. 15 (2017) 221–237. doi:10.1002/lom3.10151.

[37] L. Nanni, S. Ghidoni, S. Brahnam, Handcrafted vs. non-handcrafted features for computer vision classification, Pattern Recognit. 71 (2017) 158–172. doi:10.1016/j.patcog.2017.05.025.

[38] L. Nanni, A. Lumini, Ocean Ecosystems Plankton Classification, in: M. Hassaballah, K.M. Hosny (Eds.), Recent Adv. Comput. Vis. Theor. Appl., Springer, 2018.

[39] A.S.M. Shihavuddin, N. Gracias, R. Garcia, A.C.R. Gleason, B. Gintert, Image-based coral reef classification and thematic mapping, Remote Sens. (2013). doi:10.3390/rs5041809.